\documentclass[11pt,a4paper]{article}
\usepackage[hyperref]{ranlp2025}
\usepackage{times}
\usepackage{latexsym}

\usepackage{xurl}

\usepackage{microtype}

\aclfinalcopy % Uncomment this line for the final submission
\usepackage{graphicx} % added by author for figures

\title{%TVS 
TVS Sidekick: %\thanks{The title has been anonymised for the anonymous review.} 
Challenges and Practical Insights from Deploying Large Language Models in the Enterprise}

\author{Paula Reyero Lobo$^{1,2}$, Kevin Johnson$^{1}$, Bill Buchanan$^{1}$, Matthew Shardlow$^{2}$, \\ \textbf{Ashley Williams$^{2}$, Samuel Attwood$^{2}$} \\
$^{1}$ TVS Supply Chain Solutions, Chorley, UK \\
$^{2}$ Manchester Metropolitan University, Manchester, UK \\
\texttt{\{P.ReyeroLobo, M.Shardlow, Ashley.Williams, S.Attwood\}@mmu.ac.uk} \\
\texttt{\{kevin.johnson, bill.buchanan\}@tvsscs.com} \\
}

\date{}

\begin{document}
\maketitle
\begin{abstract}
Many enterprises are increasingly adopting Artificial Intelligence (AI) to make internal processes more competitive and efficient. In response to public concern and new regulations for the ethical and responsible use of AI, implementing AI governance frameworks could help to integrate AI within organisations and mitigate associated risks. However, the rapid technological advances and lack of shared ethical AI infrastructures creates barriers to their practical adoption in businesses.
This paper presents a real-world AI application at TVS Supply Chain Solutions, reporting on the experience developing an AI assistant underpinned by large language models and the ethical, regulatory, and sociotechnical challenges in deployment for enterprise use.
\end{abstract}

\section{Introduction}

Recent developments are driving industry interest in the field of Large Language Models (LLMs). Key developments of note are the abundant availability of commercial language modelling solutions \cite{devlin2019bert,brown2020language,thoppilan2022lamda} and the increased public awareness of the capabilities of LLMs \cite{mialon2023augmented,qu2025tool}. However, to successfully utilise these models, organisations must navigate important societal challenges related to ethics, sustainability, and compliance \cite{hagendorff2024mapping,johann2024pathways}.

%TVS SCS UK \footnote{The title has been anonymised for the double blind review.} 
%Company-X 
TVS SCS UK is a top-tier third-party logistics (3PL) provider in Europe and the UK, offering comprehensive supply chain solutions. 3PL customers increasingly adopt intelligent technology-led solutions to optimise their supply chain operations and reduce costs \cite{mehrdokht2021artificial,beibin2023large}. To stay ahead of the competition, TVS SCS UK are leveraging LLMs to create a competitive advantage and enhance their internal operational efficiency. TVS SCS UK has decided not to use third-party software integrators or product vendors for its solutions, which would negatively impact their agility and innovation. Instead, they have started their journey towards an AI transformation through an in-house AI team.

\begin{figure}[tb]
\centering
\includegraphics[width=0.5\textwidth]{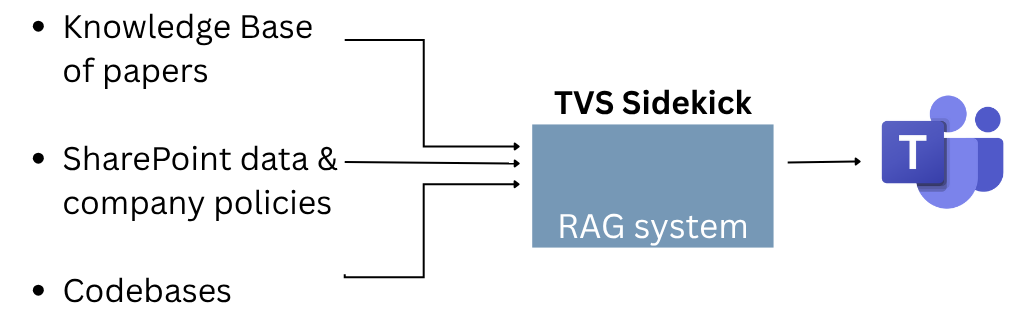}
\caption{\label{fig:overview}
Overview of TVS Sidekick, an AI assistant that leverages LLMs to answer queries with relevant enterprise data using retrieval augmented generation (RAG) via a Microsoft Teams extension.
}
\end{figure}

TVS Sidekick is the flagship product of this in-house team. TVS Sidekick %, which is fully described in Section \ref{sec:sidekick}, 
is built upon the principles of Retrieval Augmented Generation (RAG) \cite{lewis2020retrieval}. All relevant company documents, as available via their internal cloud-based systems, are vectorised and compared to the input query, with the LLM then performing information extraction for the purposes of question answering with custom prompting \cite{qu2025tool}. Users interact with TVS Sidekick via a Microsoft Teams extension (Figure \ref{fig:overview}).

As TVS SCS UK advances its AI transformation through the development of Sidekick, it must also navigate a complex legal and regulatory landscape. At the centre of this landscape are the European Union Artificial Intelligence Act (EU AIA) \cite{ec2024aia} and related standards, such as ISO/IEC 42001 for AI Management Systems \cite{iso2023standard42001}. %, both of which are described further in Section \ref{sec:case_study}
Furthermore, TVS SCS UK must overcome a range of sociotechnical challenges that accompany the deployment of LLMs, such as issues of fairness, transparency, and accountability \cite{crockett2023building,ojewale2025accountability}, which limit their practical adoption in enterprise environments.

In this paper, we report on TVS SCS UK's experience developing Sidekick, navigating the relevant legislation and regulations, and overcoming the challenges they have encountered along the way.

\subsection{Significance of this Study}

This study presents practical insights from applied AI research in a real-world business context. To be specific, we contribute to the field in three ways: 

\begin{itemize} 
    \item \textit{Technical Contributions.} We describe the design and implementation of Sidekick, an AI assistant underpinned by LLMs that is tailored for enterprise use, including novel approaches to prompt engineering and RAG.
    \item \textit{Regulatory Contributions.} We present a case study of how a business is aligning its development with emerging legislation and regulations, most notably the EU AIA, by working towards harmonised technical standards (e.g., ISO/IEC 42001).
    \item \textit{Sociotechnical Contributions.} We explore the sociotechnical challenges that accompany the deployment of LLMs in enterprise environments. We report quantitative statistics relating to the adoption of Sidekick alongside a qualitative analysis of end-user feedback.
\end{itemize}

\subsection{Structure of this Study}

The remainder of this paper is organized as follows. Section \ref{sec:related_work} reviews the relevant literature. Section \ref{sec:sidekick} details the technical implementation of Sidekick. Section \ref{sec:case_study} presents a case study of how TVS SCS UK is aligning its development with emerging legislation and regulations. Section \ref{sec:evaluation} includes a quantitative and qualitative evaluation of the progress to date. Finally, section \ref{sec:conclusion} concludes this paper and describes directions for future work.

\section{Related Work}
\label{sec:related_work}

\subsection{LLMs in the Enterprise}

% main applications
\textit{\textbf{Prominent applications, reviewing strategies to augment LLM capabilities.}} The transformer architecture enhanced language modelling capabilities and has since sparked great attention in industry \cite{vaswani2017attention}. This led to many readily available pre-trained models, which proved their superiority in fine-tuning applications \cite{devlin2019bert}. With increased data size and model complexity, decoder-only models like the generative pre-trained transformer (GPT) model series have become more attractive for industry due to their few/zero-shot performance \cite{brown2020language}. This paradigm shift led to methods for aligning to user intent \cite{ouyang2022training} (like reinforcement learning with human feedback) powering popular conversation-focused products like ChatGPT. While these scaled-up models offer business value (e.g. analysing vast data in real-time), issues such as the closed-source nature of existing solutions creates barriers to organisations lacking computational power \cite{yang2024harnessing}.

% types of rags
\textit{\textbf{Focus on approaches including RAG (and pipeline parts showing improvement).}} 
Recently, the focus has turned into giving more agency to LLMs to become independent problem solvers. For instance, by consulting with external knowledge sources for factual grounding \cite{lewis2020retrieval,thoppilan2022lamda}. More broadly, a significant step forward is the combination of ``tools'', namely tool-augmented LLMs \cite{mialon2023augmented}, including retrieval-augmented language models for efficiently handling new data. Such approaches generally consist of four stages: task planning (i.e. break down user query into tasks), tool selection, tool calling, and response generation \cite{qu2025tool}. Similarly, critical advances require frameworks for enabling LLMs to recall previous interactions \cite{zhang2024survey}, allowing for multimodal data processing \cite{sun2025survey,song2025how}, or to improve responses based on past interactions \cite{wang2024symbolic}.   

This paper presents a case study of recent LLM developments in practice, specifically through the technical implementation of an AI assistant that processes heterogeneous enterprise data sources using knowledge augmentation strategies, including novel approaches to prompt engineering and RAG.

\subsection{Responsible and Ethical AI}

% ISO is helping focus on the relevant areas.
\begin{table*}
\centering
\begin{tabular}{lr}
\hline \textbf{Principles} & \textbf{Requirements}  \\ \hline
Human oversight \& accountability & AI to support/augment humans, with humans clearly accountable. \\
Technical robustness and safety & AI tools work as expected, minimising potential harms. \\
Transparency & Clear notification of AI involvement, clear and traceable outputs. \\
Privacy \& data governance & Follow existing privacy rules with quality, robust data. \\
Diversity \& fairness & Output free of bias and does not discriminate or treat unfairly. \\
Social \& environmental wellbeing & AI is sustainable and beneficial to all. \\
\hline
\end{tabular}
\caption{\label{tab:regulatory-reqs} Key emerging principles and requirements from global AI regulations \cite{bsi2025webinar}. }
\end{table*}

%  ethical and responsible AI practices - evolving and active research area
\textit{\textbf{Challenges in training, evaluating, and deploying LLMs and emerging AI regulation.}}
While AI shows great potential and business opportunities, many concerns arise from embedding biases, contributing to climate degradation, threatening human rights and more \cite{unesco2021}. An active research area has emerged for responding hard normative questions related to AI, such as bias and fairness, transparency, and accountability \cite{jobin2019global}. Institutions at global, international, and national levels have responded with recommendations for responsible and ethical AI, consisting of principles and practices such as a human rights-centred approach to AI \cite{unesco2021}, or AI assurance methodologies (i.e. to ``measure, evaluate, and communicate the \textit{trustworthiness} of AI systems'' \cite{govuk2024}). The advent of LLMs only adds a layer of complexity to the ethical debate \cite{hagendorff2024mapping}, %raising additional concerns specific to generative AI technologies.%
raising additional concerns (regarding transparency, copyright, and safety) \cite{ec2025gpAI} that require specific regulation for generative AI technologies.

% legal stuff - EU AIA
\textit{\textbf{Global legislation and EU AIA as most far reaching and punitive of regulations}.}  %Policy makers and regulation as a way to ensure ethical AI development and use [Global - UNESCO, EU - EU AIA \& ISO/IEC, UK - AI Assurance, AI ethics research].
The EU AIA is a notable example leading the field of AI regulation, with significant non-compliance penalties to business providing or deploying AI. While legislation approaches and requirements vary across jurisdiction areas (Table \ref{tab:regulatory-reqs}), AI regulations are developing globally to provide assurances in critical aspects such as human oversight and accountability, technical robustness and safety, or privacy and data governance \cite{bsi2025webinar}. 

% regulations - ISO42001 and link to EU AIA
%Key emerging themes \& requirements (Table \ref{tab:regulatory-reqs}), highlighting ISO 42001 for enterprise.
Governments and legislative bodies are working towards practical strategies to implement the principles underlying AI regulations. %, largely through the development of harmonised standards \cite{standards2024webinar}. 
Harmonised standards are one of the primary mechanisms for helping organisations translate regulatory requirements into technical implementations \cite{standards2024webinar}. 
Standardisation should specify minimum technical testing, documentation, and public reporting to limit AI developers and/or users discretion in complying with regulatory requirements \cite{johann2024pathways}. However, local empirical studies and specific examples of how organisations implement processes that ensure AI regulation principles \cite{brenner2024professionals} is crucial for a democratic approach to ethical and responsible AI.

% broader considerations - public involvement / co-creation aluded to in ISO42001?
\textit{\textbf{From theory to practice.}} %Relevant examples of how to apply AI governance \& regulation [co-production, inquiries - Policy Connect, prominent research - toolkit evaluation].
While approaches to ethical AI exist (including bias tests, %toolkits, 
checklists and risk impact assessments), organisations face barriers that limit their practical adoption \cite{crockett2023building}. Technical approaches alone are not sufficient to establish an ethical AI infrastructure \cite{ojewale2025accountability}. Instead, participatory approaches involving civil society stakeholders are needed for effective standard setting, implementation, and enforcement \cite{crockett2024peas,modhvadia2025how}. 
This paper contributes to bridging the gap between theory and practice through the experience of implementing an AI governance strategy in a real-world business context, reporting on the technical, legal and human challenges involved with the adoption of generative AI technologies.

\subsection{Positioning this Study}

%Maybe describe how AI/LLMs sparked attention in the logistics sector in particular but challenges for effective and safe use.
In the logistics sector, real-time data analysis can transform business operations, from internal warehousing and inventory processes to stakeholder management \cite{mehrdokht2021artificial}. However, empirical research in related areas \cite{qian2024understanding,kapania2025examining} shows that benefits and trade-offs in the use of AI technologies manifest differently depending on their application domain. % or context.
%However, as the complexity of AI systems increases, their decision-making processes become difficult to trace. To businesses this translates as the urgent need to have an effective AI strategy and governance in place for understanding and mitigating associated risks while continually improving AI systems \cite{iso2023standard42001}.

%The key contribution of the paper: on empirical experience of technical deployment of LLMs and AI management system (AIMS) implementation at TVS SCS UK.

Despite growing understanding of public attitudes towards AI \cite{modhvadia2025how,mhasakar2025perspectives}, research on its industrial application remains limited. This study presents %practical 
insights from the development and use of LLMs at TVS SCS UK, %. Specifically, 
to address the following gaps: 
\begin{itemize}
    \item  Examining the implementation and practical application of recent LLM advances within the enterprise context. 
    \item Embedding high-level ethical principles in AI regulatory frameworks into organisational practices. 
    \item  Empirical analysis of challenges that emerge with adopting LLMs in a logistics company. %the adoption of LLMs in industry. %[the logistics sector.] 
\end{itemize}

\begin{figure*}[htb]
\centering
\includegraphics[width=0.9\textwidth]{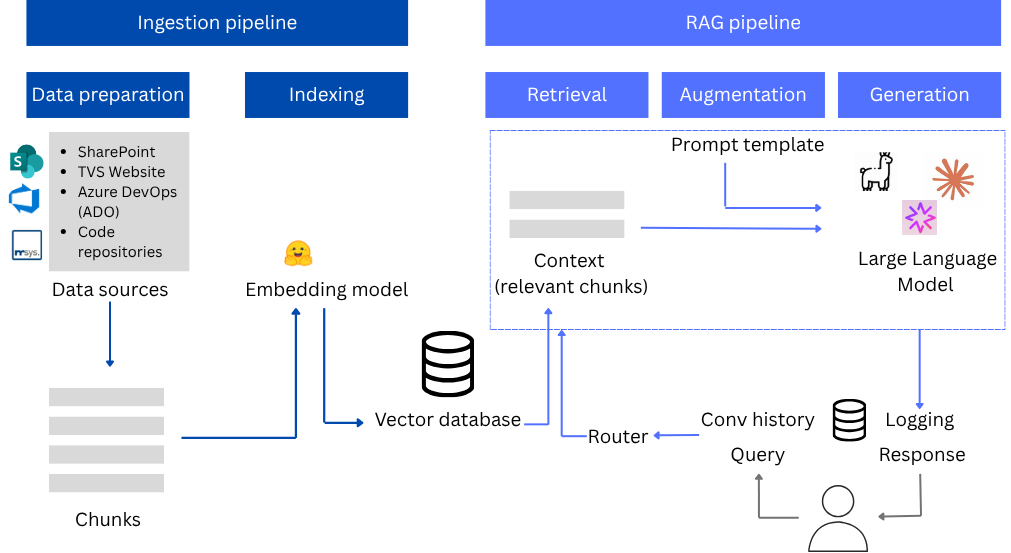}
\caption{\label{fig:tech-arch}
Architecture diagram showing the main components of Sidekick, namely the ingestion and RAG pipelines, with novel approaches to prompt engineering (to handle code queries) and augmentation retrieval (for tool use).%, which are described in detail in Section \ref{sec:sidekick}.
}
\end{figure*}

\section{Technical Implementation}
\label{sec:sidekick}

This section presents the design and implementation of Sidekick (Figure \ref{fig:tech-arch}), describing: (i) the integration of relevant company data into a vector database (Ingestion pipeline), and (ii) how this vectorised data is used to process user queries with enhanced LLM capabilities (RAG pipeline). 

\subsection{Ingestion Pipeline}

Vector databases are increasingly used to enhance LLM-generated outputs by providing relevant text fragments (``chunks'') that have a similar meaning to the user query (i.e. ``context''). To do so, company data needs to be transformed and embedded into a common database that handles semantic similarity searches. The \textit{vector database} acts as a bridge between the two system components, accelerating the retrieval of content that is relevant to the user query. 

%To address this challenge, the implemented solution is based on the following steps:
The first system component integrates information from different company data sources into the vector database, in two main steps:
% high-level overview description of what happens in each step 
\begin{itemize}
    \item Data preparation. First, TVS data is fetched from different \textit{data sources}, i.e. SharePoint, Azure DevOps (ADO), code repositories, and TVS website, with a scheduled hour refresh. % %[Table/description of data from each source (i.e. lists, site, drives from SP, ADO task items, and the RPG repository)].  
    Data is then processed to extract \textit{chunks} using a document loader: i.e. parsing (extract or transform to text - for code) and chunking (splitting by semantic or logical boundaries). 
    \item Indexing. Extracting semantic vectors from each chunk with an \textit{embedding model}, and creating an index in the vector database for each data source (to define specific fields). %Specifically, using Hugging Face sentence transformers and Elasticsearch.
\end{itemize}

% Implementation (highlighting novel approaches)
%To integrate different and scale to new data sources, Sidekick fetches for each source and uses a common document loader, with specific index for each source (different fields of interest for each source)

Sidekick is developed to handle both text and code-related queries. Crucially, %with the following approach for code integration
using prompt engineering for code integration. First, files are split to objects by logical meaning (i.e. functions, methods, or procedures). An LLM is prompted to generate descriptions to each code file, using its object list to report on the overall purpose, structure, key procedures, functions, and external interactions. Both code and transformed text fragments are stored in the vector database, to expose relevant source code lines as sources when responding to the user query.% to address LLM hallucinations. 

%splitting files by their logical meaning (i.e. functions To integrate code source, data is [logical splitting and transformed to text by ...]. Furthermore, raw code and transformed text fragments are stored [text fragment to retrieve and generate response, combining raw files].

\subsection{RAG Pipeline}

The second system component processes user queries by leveraging company data and conversation history to enhance LLM outputs. %Sidekick applies RAG principles and novel approaches including prompting techniques.

The user query and conversation history (i.e. queries and responses of the last 60-minute session) are sent to a \textit{router}. The router splits the user query into sub-sentences (i.e. specific tasks) and calls an LLM to decide which route to take for the augmentation retrieval. Each route uses a type of ``chatterbot'', a tool-based LLM optimised to answer questions related to different data sources.

Each task identified from the user query triggers an instance of RAG:
\begin{itemize}
    \item Retrieval: information retrieval from vector database using the same embedding model to extract \textit{context} (top-10 similar chunks) and reformat chunks (its text and metadata as XML or JSON list for code route).
    \item Augmentation: calls an LLM to extract the required parameters to generate the answer (including prompt template).
    \item Generation: calls an LLM using the instructions and context from previous steps.
\end{itemize}

The output generated for each task are combined into a single \textit{response} using the LLM only with generated texts. The user query and response are saved for logging and leveraging conversation history. %into the conversation history, as well as for logging purposes.

%\section{Navigating Regulatory Challenges of LLM Deployment: Case Study}
\section{Navigating Regulatory Challenges of TVS Sidekick: Case Study}
\label{sec:case_study}

This section presents the regulatory challenges that emerge with the development of LLMs, and how they may be overcome in a real-world business context. Specifically, we present a case study on navigating a complex and changing AI regulatory landscape in the enterprise, leading to the implementation of the first harmonised technical standard for responsible AI development and use. 

\subsection{EU AIA \& Harmonised Standards}
% how legislation landscape was navigated in practice, leading to identification of ISO/IEC 42001

% horizontal standardisation request by the European Commission: https://www.iso.org/sectors/it-technologies/ai
\begin{table}
\centering
\begin{tabular}{lr}
\hline \textbf{Requests} & \textbf{Standard(s)}  \\ \hline
Accuracy & 23282* \\
Robustness & 24027, 12791 \\
Transparency & 12792* \\
Human oversight & 8200, 42105* \\
Data and Data Management & 25012, 5259 \\
Cybersecurity & 27001 \\
Record keeping and logging & 24970* \\
Quality management systems & 9001, 25059* \\
Risk management systems & 31000, 23894 \\
%& 42005 \\
Conformity assessment & 42006 \\
\hline
\end{tabular}
\caption{\label{tab:cen-cenelec} Horizontal standardisation request for the EU AIA \cite{standards2024webinar}, mapped to available ISO/IEC standards. Highlighted standards (*) are yet to be published (20th August 2025).}
\end{table}

%The EU AIA [how it was identified as important for TVS.]
TVS SCS UK is achieving compliance working towards AI standardisation, which is key to the development and adoption of AI.  One key regulation shaping the field of standardisation is the EU AIA, which is leading the global landscape of AI regulation.

Different harmonised standards are being developed to support the implementation of the EU AIA, such as the ISO/IEC 12792 and 24970 standards for addressing the transparency and logging of AI systems, respectively (see Table \ref{tab:cen-cenelec}). Building upon relevant standards, including AI Concepts and Terminology (22989) and AI Risk Management (23894), ISO/IEC 42001 is the first international standard for AI Management Systems, aiming to guide organisations in the responsible development and use of AI systems. 

Recognising the value of standards to operationalise AI regulation principles for ethical and responsible AI, TVS SCS UK has decided to adopt an AI Management System (AIMS) framework to develop trustworthy AI solutions.

\subsection{ISO/IEC 42001 Implementation} \label{subsec:iso_imp}
% How they plan to implement the standard, with the gap analysis and processes (including items in the work plan for the gaps)
TVS SCS UK have developed and deployed formal management systems in important areas such as information security, quality, health and safety, business continuity, and environmental management. To effectively implement an AI management system, TVS SCS UK began with mapping the key requirements of ISO/IEC 42001 to existing standards, focusing on management systems already adopted by the organisation. 

\begin{table}
\centering
\begin{tabular}{llc}
\hline \textbf{42001} & \textbf{Requirement} & \textbf{Focus} \\ \hline 
%4.1, 4.3 & Scope and context &  &  \\ % Policy document: Purpose
%4.2 & Needs and expectations & Y &   \\ % Policy document: Requirements
4.[1/2/3] & Purpose \& Requirements  \\
%6.2 & AI Objectives &  \\ % Policy document: Objectives
%6.3 & Change management &    \\ % Policy document: Change 
6.[2/3] & Objectives \& Change & \\ % Policy document
%5.1 & Leadership commitment &   \\ % Policy document: Leadership
%5.2 & AI policy &   \\ % Policy document: Policy
5.[1/2] & Leadership \& Policy & \\ % Policy document
5.3 & Roles \& responsibilities &     \\ % Policy document: Accountability
%5.3 & Accountability &     \\
%6.1.[1/2/3],  & Risks \& Impact & Y    \\ % Audit, Impact assessment
6.1.[1/2/3],  & AI Risks & Y    \\
8.[1/2/3/4] & & \\
9.1. 9.2.[1/2] & Monitoring \& Measuring & Y  \\ % Monitoring, 
10.[1/2], 9.3 & Continuous improvement & Y  \\ % Resources, Management review
7.[1/2/3/4],  & Awareness \& Training &   \\% Resources, Competence, Tech docu
7.5.[1/2/3] & & \\
%9.3 & Management review & & \\
\hline
\end{tabular}
\caption{\label{tab:mapping} %Mapping analysis between the first AI Management System (ISO/IEC 42001) and Information Security (ISO/IEC 27001) standards. 
Mapping analysis between ISO/IEC 42001 and existing management systems at TVS SCS UK, highlighting focus areas for implementation (``Y'').}
\end{table}

The results from this mapping analysis are shown in Table \ref{tab:mapping}. Notably, TVS SCS UK maintains an Information Security management system following ISO/IEC 27001 \cite{iso2022standard27001}. Processes supporting this standard, especially related to data management and cybersecurity, %are closely aligned to AIMS framework requirements of ISO/IEC 42001
were aligned with ISO/IEC 42001 requirements. This comparison helped to identify focus areas for developing an AIMS:%AI quality and risk management systems. % or: to start developing an AI quality and risk management system

\textit{\textbf{AI Risks}}. %TVS SCS UK maintains a risk management strategy to periodically assess risks, which is an integral part of their information security. 
TVS SCS UK maintains a risk management strategy as an integral part of their information security. This ongoing process sets out responsibilities and a methodology to periodically assess risks based on likelihood and impact levels. One of the main challenges introducing AI is the need of staying relevant with current risks. To this end, TVS SCS UK is working towards establishing a \textit{Knowledge Base} that informs AI development and use within the company. The AI team started maintaining an academic database of research reviews including meta-analyses and relevant case studies that is accessible throughout the company; both in full-text and via their in-house AI assistant for the purpose of question answering. Furthermore, a systematic search \cite{brereton2007systematic} of AI research papers in relevant topics (Table \ref{tab:kb}) allows to explore topic distribution and relevant metadata, such as indexed keywords or keywords from the authors, and supports the maintenance and updating of the academic database. 

%(ii) automatic metadata analysis (topic distribution [ add table of topics under Performance, Safety, Regulation categories], keyword extraction / plot, and manual analysis feeding into academic].

\begin{table}[tb]
    \centering
    \begin{tabular}{lr} 
    \hline
    Category & Topics \\ %\hline
    Performance     &  alignment, reliability, robustness,  \\
    & prompt engineering, usefulness, \\
    & helpfulness, truthfulness \\
    Safety     & privacy, security, safety \\
    & interpretability, transparency, \\
    & explainability, fairness, trust- \\
    & worthiness, adversarial attacks \\
    %& attacks, human-in-the-loop \\
    Regulation & regulation, best practice*, gover- \\
    & nance, compliance, accountability  \\ \hline
    \end{tabular}
    \caption{Topics of AI/LLM performance, safety, and regulation feeding into the \textit{Knowledge Base} of papers.}
    \label{tab:kb}
\end{table}

\textit{\textbf{Monitoring \& Measuring}}. Another component of the AIMS framework is to capture monitors and measures on the use of AI, including an internal audit programme. TVS SCS UK is developing a \textit{monitoring system} supporting Sidekick, which includes usage indicators %(i.e. interaction volume, response time, user engagement) 
 (Table \ref{tab:monitoring}) and descriptive metrics of interactions (volume breakdown by department, job title, individual user, and question type). Ultimately, these metrics aim to pragmatically measure the effectiveness of the AI assistant, setting a starting point for other AI performance and safety measures. For instance, obtained through the provision of feedback channels \cite{helma2024challenges} to report quality or safety incidents, or the inclusion of LLM observability evaluations \cite{kenthapadi2024grounding}. 
 
 %observability evaluations/%Draw attention in the development of performance evaluation, internal audits, and management review process. Monitoring system to track AI usage. Future work/conclusion: Feedback loops to report quality incidents & flag safety incidents + internal audit (dev process not only AI outputs).
\begin{table}[tb]
    \centering
    \begin{tabular}{l|l} \hline
    Usage indicators     &  \\
    Interaction volume & Number of messages (i.e. \\
    & prompts) and unique users.\\
    Response time & Average response time (s). \\
    %& in seconds. \\
    User engagement     & Average of messages per \\
    & session (on daily basis). \\ \hline
    %Descriptive metrics & 
    \end{tabular}
    %\caption{Monitors on the use of the AI assistant at TVS SCS UK.}
    \caption{Description of metrics in the \textit{monitoring system} supporting the AI assitant at TVS SCS UK.}
    \label{tab:monitoring}
\end{table}

\textit{\textbf{Continuous improvement.}} The effective management of vulnerabilities to the AIMS is crucial for demonstrating continual improvement in the use of AI, with documented validation and verification. TVS SCS UK is establishing processes for maintaining and deploying AI, primarily focused on the evaluation and technical documentation of Sidekick. To this end, a primary evaluation objective has been set to understand the needs and ways in which the AI assistant may best support different company roles and responsibilities. Specifically, through the organisation of periodic \textit{feedback interviews} as part of a continuous evaluation of Sidekick, with target populations whose adoption of AI could bring most benefit to the company. A participatory approach to AI development aims to support a culture of ethical and responsible AI. 

%Challenging given the open nature + ultimately understand barriers in deployment, so \textit{feedback interviews} with target population (i.e. adopting AI could bring most benefit to the company). % Future work/conclusion: Scaling up with surveys and apointing AI Champions to meet requirements & foster a responsible AI culture.
%Demonstrate continuous evaluation and improvement of the AIMS. Continuous evaluation and interviews with target population. AI Champions \& culture. 

% v2 is with size 0.72
\begin{figure*}[htb]
\centering
\includegraphics[width=0.7\textwidth]{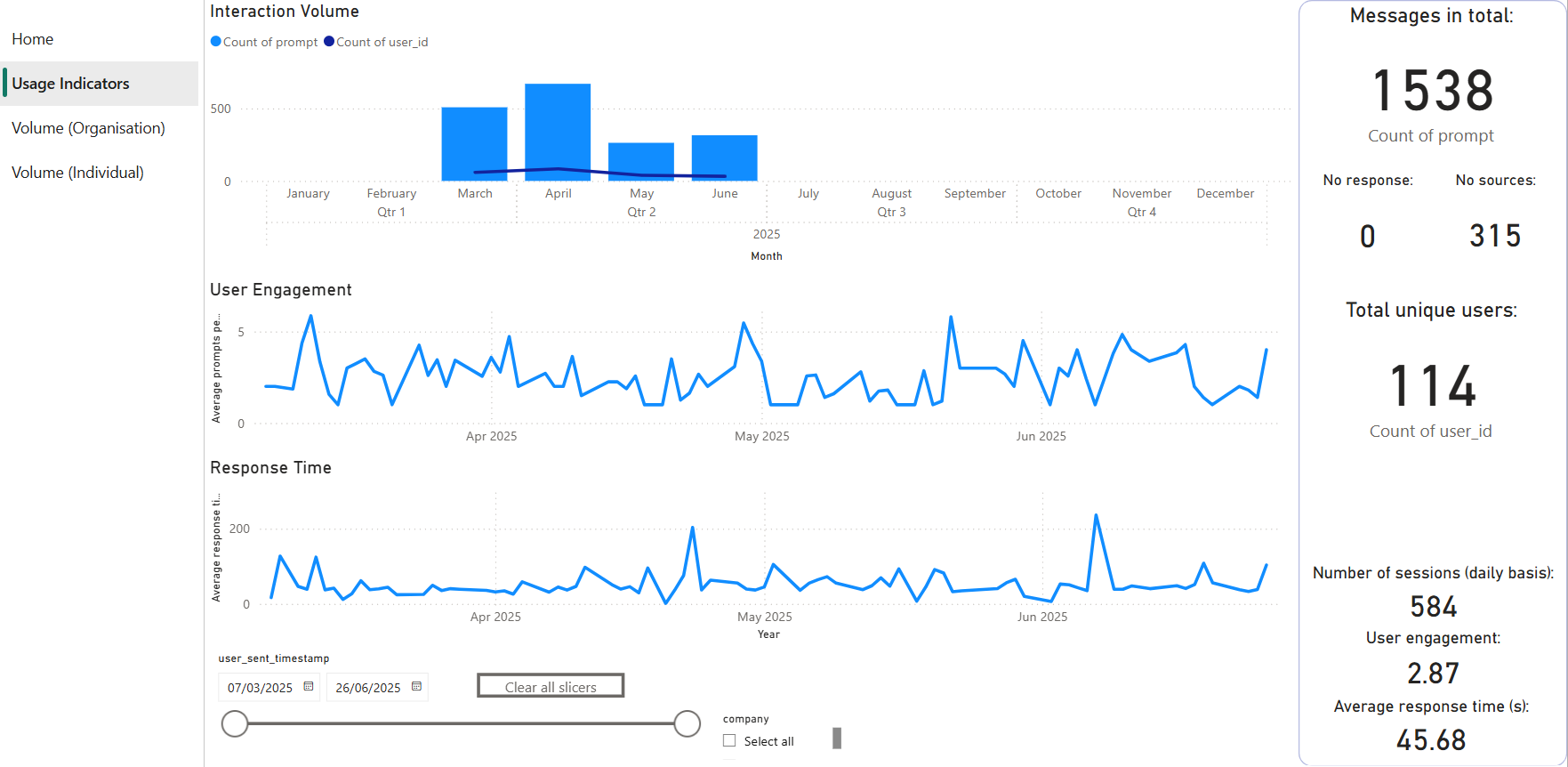}
\includegraphics[width=0.7\textwidth]{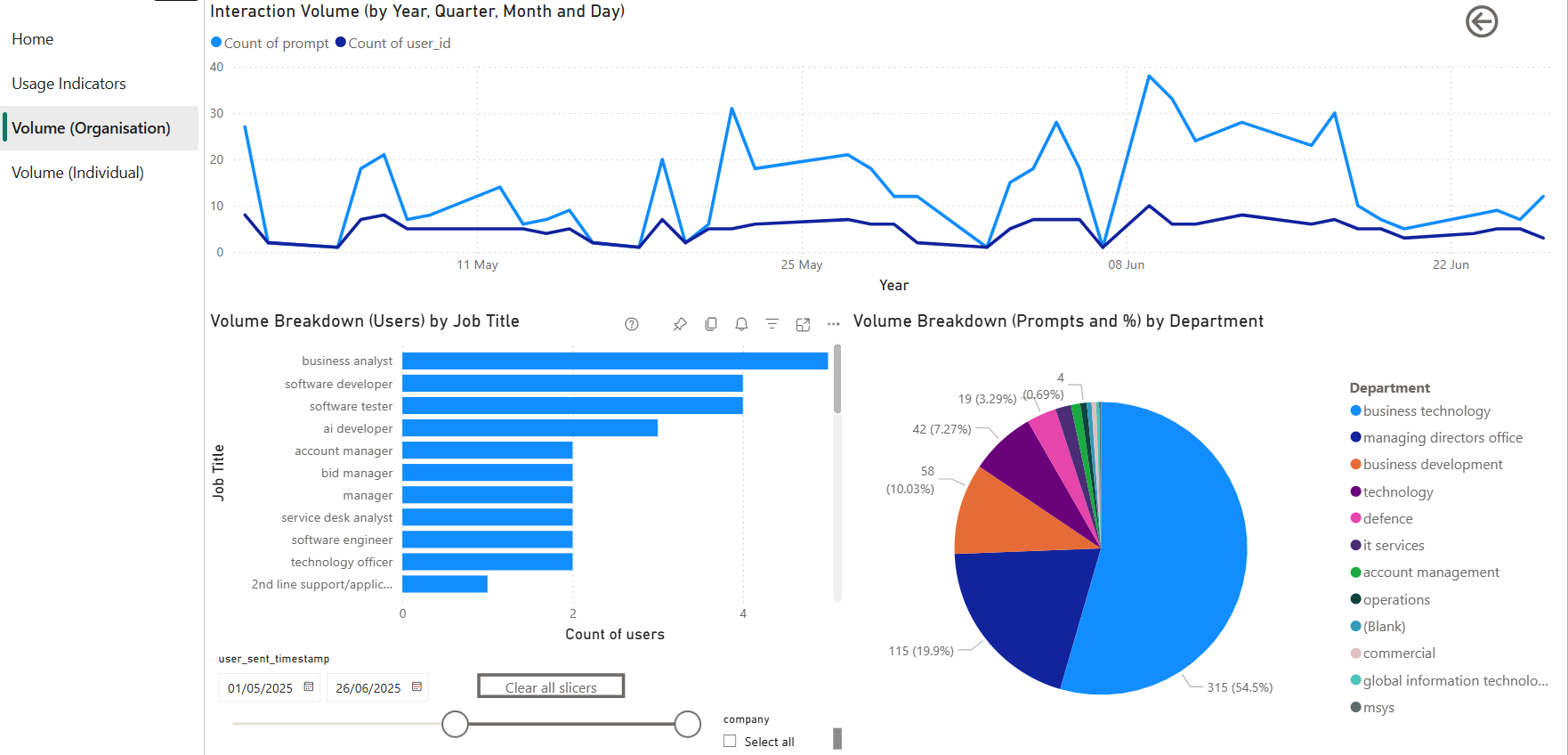}
\includegraphics[width=0.7\textwidth]{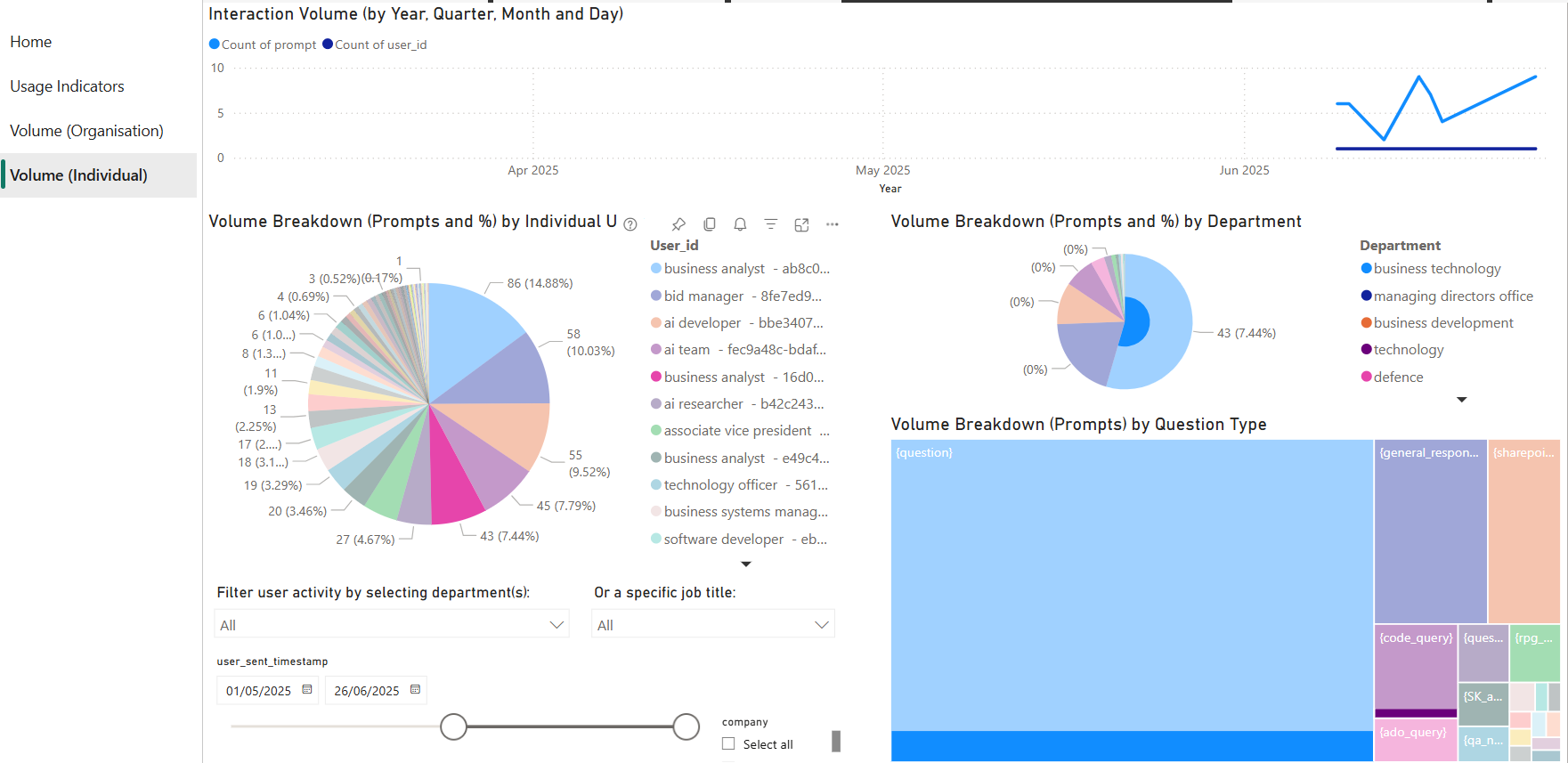}
\caption{\label{fig:dashboard}
%Monitoring system to evaluate the effectiveness of the AIMS based on real-time usage data.
Monitoring system measuring real-time usage data of TVS Sidekick.
}
\end{figure*}

\section{Monitoring \& Evaluation}
\label{sec:evaluation}
% Practical insights on the deployment of LLMs

This section presents insights gathered from the deployment of LLMs at TVS SCS UK, highlighting sociotechnical challenges in %their application to professional environments. Specifically, we present empirical findings from the monitoring system and initial evaluation of the sidekick product.
their enterprise use. Following the on-going implementation of an AI governance model, we specifically report on empirical findings from the monitoring system and initial evaluation of the Sidekick product.

\subsection{Adoption \& Usage}

The implementation of an AIMS framework following ISO/IEC 42001, in particular related to \textit{Monitoring \& Measuring} requirements, provides practical insights on the levels of AI adoption and usage in the organisation. Consequently, we report findings from the monitoring system described in Section \ref{subsec:iso_imp}.

Figure \ref{fig:dashboard} shows quantitative statistics related to the initial adoption of Sidekick at TVS SCS UK. The monitoring system shows usage indicators and descriptive metrics of interaction volume within a 4-month period (March-June 2025).

%[Results: adoption rates - which departments, usage metrics, discuss figure with dashboard.] 
%Adoption rates based on metrics: volume, retention, engagement, response time.
Overall, continued use of the AI assistant is shown within the observed time period. This is seen in continued measures both in terms of user engagement and interaction volume (exceeding 500 prompts in the first two months and 250 in the following two months). Despite fluctuations, the conversations do not seem to be long, rarely exceeding an average of five questions per conversation. The response time has peaks on specific dates that increment the time to 46 seconds on average.

The descriptive analysis of interaction volume at organisational level reveals that the most active users were primarily in technology (e.g. developers) and business roles (e.g. bid management, business analysts). At the departmental level, these roles correspond to IT (business technology, business development), management, and operational areas such as defence, technology, commercial, and operations.  %[Job titles: IT (developers) and business (bid management, business analyst.] [Correspond to IT-related departments (business technology, business development, managing directors office) and use cases (defence and technology). ]

In terms of individual usage, the breakdown of activity per individual makes a clear distinction between lead and early adopters (i.e. 46 - 253 queries) and occasional users (less than 20 on average). Queries answered with SharePoint data (i.e. \textit{question}) were the most common, followed by responses without retrieval augmentation (\textit{general\_response}), codebase file queries (\textit{rpg\_query}) and queries related the development environment, i.e. Azure DevOps (\textit{ado\_query}). 
% [lead and early adopters vs occasional users]. [Mostly related to SharePoint (general questions and general response - i.e. no RAG) followed by codebase related queries, with lower usage rates on ADO queries.]

\subsection{Qualitative Feedback}
% SA note: bid management experience? RPG developer experience?

\begin{table}[tb]
    \centering
    \begin{tabular}{ll}
    \multicolumn{2}{l}{Understanding current use of AI/Sidekick}  \\ \hline
         %& Have you used Sidekick/other AI tools? \\
         & Have you used Sidekick/other AI tools? \\
         %& tools? \\
         & What have you used it for? \\
         & Where was AI/Sidekick most helpful/\\
         & unhelpful? \\
    \multicolumn{2}{l}{Outlook}\\ \hline
        & In what aspects of your job would AI\\
        & be most useful? \\
        & Do you have any concerns about inte-  \\
        & grating AI into your workflow? \\
    \end{tabular}
    \caption{Topic guide of \textit{feedback interviews} supporting the continuous evaluation of TVS Sidekick.}
    \label{tab:interviews}
\end{table}

The initial round of feedback interviews that feed into the \textit{Continuous improvement} requirement under ISO/IEC 42001 highlights significant challenges when introducing AI in the business context. Primarily, with respect to the perceived benefits and risks of deploying LLMs in the enterprise, due to Sidekick being the flagship product. 

In total, 24 interviews with members of the IT department at TVS SCS UK were conducted between March and April 2025. Participants were invited to 30-minute online meetings for a semi-structured interview. The topic guide (Table \ref{tab:interviews}). included questions i) to gather experiences so far in using AI/Sidekick at work and ii) understand how TVS staff want to use Sidekick in the future. Finally, interview minutes were thematically analysed \cite{byrne2022worked} by two independent coders. 

The analysis of qualitative feedback led to better understanding of baseline attitudes towards AI.  
The following themes were identified:
% \begin{itemize}
%     \item Enhanced retrieval (Mentioned by: 9):
%     \item Good for business analysts/Makes job more enjoyable (Mentioned by: 11)
%     \item Doubts around accuracy of sources (Mentioned by: 7)
%     \item Good extracting business logic (Mentioned by: 5)
%     \item Not enough technical detail (Mentioned by: 9)
%     \item Enough technical detail (Mentioned by: 4)
%     \item Keen to engage with AI/Proposing new features (Mentioned by: 8)
%     \item Privacy/commercially sensitive questions (Mentioned by: 5)
%     \item Concerns/reluctance/skepticism ((Mentioned by: 3)
% \end{itemize}

\textbf{\textit{Enhanced retrieval}} (Mentioned by: 16). A key advantage of Sidekick over other tools is its specificity to TVS data. Users valued its assistance with SharePoint-related tasks, finding it faster than a manual search and with a ``readable and visible'' format, especially for the source list.

\textit{\textbf{Good extracting business logic}} (Mentioned by: 10). Sidekick was particularly helpful in providing business knowledge, with clear use cases for business analysts. Specifically, for understanding the context of TVS data and key definitions of components within business processes.

\textit{\textbf{Not enough technical detail}} (Mentioned by: 13). Developers emphasized the need for more domain knowledge to explain internal programmes. Particularly, those relying on a legacy programming language with limited technical documentation. The current version of the AI assistant offers a good starting point for understanding key parameters and functions, but remains limited in addressing more specific queries from technical users.

\textit{\textbf{Keen to engage with AI}} (Mentioned by: 11). Overall, staff were enthusiastic about using Sidekick to standardise code, reduce duplication, refer new starters to source documentation, or avoid ownership issues when using external AI tools. Furthermore, new features were proposed, including learning from user prompts or returning questions to users to resolve ambiguous queries. 

\textit{\textbf{Privacy/commercially sensitive questions/Other concerns}} (Mentioned by: 8). There were no major concerns with the use of Sidekick, provided it was fed with the right information and access levels. Concerns were raised around job security and distrust in AI tools, along with the emphasis on using Sidekick internally due to potential disclosure of information from the client side.
%Privacy/commercially sensitive questions (Mentioned by: 5).

The first round of feedback has been worked into a plan for continual improvement and addressing concerns, informing further developments of TVS Sidekick. TVS SCS UK will continue developing processes to adhere to ethical principles in regulatory standards, sharing practical insights in critical areas such as managing AI risks, providing relevant monitors and measures on AI use, and increasing AI adoption through training and consultation.

% \section{Conclusion \& Outlook}
\section{Conclusion}
\label{sec:conclusion}

This paper presented the experience and challenges encountered in a real-world business scenario with the development and deployment of TVS Sidekick, an AI assistant leveraging LLMs for enterprise use. This empirical study provides practical knowledge, including key lessons learned from the implementation and governance of the in-house AI assistant.

\section*{Limitations \& Ethical Considerations}

The findings and insights presented are drawn from a specific organisational context and reflect experiences within a particular time frame and initial phase of evaluation. While the technical specifics and detailed implementation of each component of the governance framework are outside the scope of this work, this paper aims to contribute to the wider community by sharing reflections on navigating technical, ethical, regulatory, and sociotechnical challenges of deploying LLMs in practice.

% \section*{Acknowledgments}
% This project was co-funded by UKRI through Innovate UK and TVS Supply Chain Solutions.
% The acknowledgments should go immediately before the references. Do not number the acknowledgments section.
% \textbf{Do not include this section when submitting your paper for review.}

\bibliographystyle{acl_natbib}
\bibliography{anthology,ranlp2025,bib}

%\appendix

\end{document}